\RequirePackage{fix-cm}
\documentclass[twocolumn]{svjour3}          % twocolumn
\smartqed  % flush right qed marks, e.g. at end of proof

\usepackage[utf8]{inputenc} % allow utf-8 input
\usepackage[T1]{fontenc}    % use 8-bit T1 fonts
\usepackage{url}            % simple URL typesetting
\usepackage{booktabs}       % professional-quality tables
\usepackage{amsfonts}       % blackboard math symbols
\usepackage{nicefrac}       % compact symbols for 1/2, etc.
\usepackage{microtype}      % microtypography
\usepackage{xcolor}         % colors
\usepackage{graphicx}
\usepackage{amsmath}
\usepackage{amssymb}
\usepackage{multirow}
\usepackage{color, colortbl}
\usepackage{subcaption}
\usepackage{subfloat}
\usepackage{tabularx}
\usepackage[export]{adjustbox}
\usepackage{threeparttable}
\usepackage{pifont}
\usepackage[misc]{ifsym} % use \Letter
\usepackage{algorithm}
\usepackage{algorithmic}
\usepackage{paralist}
\usepackage{listings} 
\usepackage{amssymb}
\usepackage{bbding}
\usepackage[authoryear]{natbib} 
\setcitestyle{numbers,square} %for \citet
\usepackage{graphicx}

\definecolor{COLOR_CSID}{HTML}{e0f5ff}
\definecolor{COLOR_NEAROOD}{HTML}{ffefe0}
\definecolor{COLOR_FAROOD}{HTML}{ffdebf}
\definecolor{COLOR_MEAN}{HTML}{f0f0f0}

\usepackage{xspace}
\usepackage{comment}
\usepackage{bm}
\usepackage{wrapfig}
\usepackage{xcolor}
\usepackage{colortbl}

\usepackage[colorlinks,linkcolor=red,anchorcolor=blue,citecolor=blue,CJKbookmarks=True]{hyperref}

\definecolor{citecolor}{HTML}{0071BC}
\definecolor{linkcolor}{HTML}{ED1C24}

\makeatletter
\renewcommand\paragraph{
  \@startsection{paragraph} % name
  {4} % level
  {\z@} % indent
  {.5em \@plus1ex \@minus.2ex} % beforeskip
  {-1.5em} % afterskip
  {\normalfont\normalsize\bfseries} % style
}
\DeclareRobustCommand\onedot{\futurelet\@let@token\@onedot}
\def\@onedot{\ifx\@let@token.\else.\null\fi\xspace}

\makeatother

\begin{document}

\sloppy 

% \title{SFNet: Faster, Accurate, and Domain Agnostic Semantic Segmentation via Semantic Flow}

\title{Sample Correlation for Fingerprinting Deep Face Recognition}
%\titlerunning{Short form of title} % if too long for running head
% (\textrm{\Letter})
\author{Jiyang Guan$^{1,2}$ \and
        Jian Liang$^{1,2}$\and
        Yanbo Wang$^{1,2}$ \and
        Ran He$^{1,2}$
}
        
\institute{
    Jiyang Guan (guanjiyang2020@ia.ac.cn) \\ 
    Jian Liang (liangjian92@gmail.com) \\
    Yanbo Wang (wangyanbo2023@ia.ac.cn) \\
    Ran He (rhe@nlpr.ia.ac.cn) \\
    1 MAIS $\&$ CRIPAC, Institute of Automation, Chinese Academy of Sciences \\
    2 School of Artificial Intelligence, University of Chinese Academy of Sciences
    }
\date{Received: date / Accepted: date}


\maketitle

\begin{abstract}
Face recognition has witnessed remarkable advancements in recent years, thanks to the development of deep learning techniques.
However, an off-the-shelf face recognition model as a commercial service could be stolen by model stealing attacks, posing great threats to the rights of the model owner.
Model fingerprinting, as a model stealing detection method, aims to verify whether a suspect model is stolen from the victim model, gaining more and more attention nowadays.
Previous methods always utilize transferable adversarial examples as the model fingerprint, but this method is known to be sensitive to adversarial defense and transfer learning techniques.
To address this issue, we consider the pairwise relationship between samples instead and propose a novel yet simple model stealing detection method based on SAmple Correlation (SAC).
Specifically, we present SAC-JC that selects JPEG compressed samples as model inputs and calculates the correlation matrix among their model outputs.
Extensive results validate that SAC successfully defends against various model stealing attacks in deep face recognition, encompassing face verification and face emotion recognition, exhibiting the highest performance in terms of AUC, p-value and F1 score.
Furthermore, we extend our evaluation of SAC-JC to object recognition datasets including Tiny-ImageNet and CIFAR10, which also demonstrates the superior performance of SAC-JC to previous methods.
The code will be available at \url{https://github.com/guanjiyang/SAC_JC}.
\keywords{Model Fingerprinting \and Deep Face Recognition}
\end{abstract}
\section{Introduction} \label{sec:intro}

% In recent years, Deep Neural Networks (DNNs) have achieved impressive performance in a variety of tasks, including but not limited to face recognition \cite{taigman2014deepface}, medical diagnosis \cite{kermany2018identifying}, and autonomous driving \cite{tian2018deeptest}.
% Deep face recognition, as an important implementation of DNNs, has achieved great success.
% Face recognition \cite{wang2021deep} has witnessed remarkable advancements in recent years, thanks to the development of deep learning techniques.
In recent years, remarkable advancements in face recognition have been largely attributable to the development of deep learning techniques \cite{wang2021deep}.
A common practice for model owners is to offer their models to clients through either cloud-based services or client-side software.
Generally, training deep neural networks, especially deep face recognition models, is both resource-intensive and financially burdensome, requiring extensive data collection and significant computational resources. 
Therefore, well-trained models possess valuable intellectual property and necessitate protection \cite{jia2021entangled, he2019sensitive}.
Nonetheless, model stealing attacks can steal these well-trained models and evade the model owners' detection with only API access to the models \cite{lukas2020deep}, posing serious threats to the model owner's Intellectual Property (IP).

Model stealing attacks are carried out with the goal of illegally obtaining functionally equivalent copies of the well-trained model owners' source model, with the white-box or even the black-box access to the source models.
In the case of white-box access, the attacker can gain access to all the internal parameters of the source model. 
To avoid detection by the model owner, the attacker is able to employ source model modification, including pruning \cite{liu2018fine}, fine-tuning \cite{molchanov2019importance}, adversarial training \cite{shafahi2019adversarial}, and knowledge distillation \cite{huang2022evaluation}.
Furthermore, attackers are also able to leverage the model extraction attack \cite{jagielski2020high,orekondy2019knockoff} to steal the function of the source model, with only the black-box access to the source model.
In such a paradigm of model stealing attack, the attacker can steal the function of the source model using only the model's outputs, without the need for access to the inner parameters, and thus is considered more general and threatening.
Regarding deep face recognition models, we observe that model extraction attacks achieve an accuracy of up to $95.0\%$ of the original accuracy of the source model on KDEF \cite{goeleven2008karolinska} in face emotion recognition with only output labels of the source model. 
Moreover, in face verification, we also observe attackers can evade most of the stealing detection easily because these models only output the verification results rather than labels.
In total, deep face recognition is confronted with a pressing challenge posed by model stealing attacks.

In recent years, the growing concerns over model stealing attacks have led to the development of various methods aiming at protecting the intellectual property (IP) of the deep models. 
Generally, these methods can be categorized into two categories: the watermarking methods \cite{uchida2017embedding,chen2018deepmarks,fan2019rethinking,zhang2020passport,adi2018turning,zhang2018protecting,fan2021deepip,jia2021entangled,ge2021anti} and the fingerprinting methods \cite{lukas2020deep,cao2021ipguard,li2021modeldiff,peng2022fingerprinting,wang2021fingerprinting}.
Watermarking techniques typically incorporate either weight regularization methods \cite{uchida2017embedding,chen2018deepmarks,fan2019rethinking} or backdoor insertion strategies \cite{adi2018turning,zhang2018protecting,jia2021entangled} during the model training phase to embed a distinct watermark into the model. 
However, these approaches need to manipulate the model's training process, often resulting in a trade-off where the model's performance on its main task is compromised. 
On the contrary, fingerprinting methods leverage the transferability of adversarial examples and identify stolen models by calculating the attack success rate on the suspected model.
These methods do not interfere with the model's training procedure, which means they do not sacrifice the model's accuracy on its main task.
However, it's important to note that adversarial-example-based fingerprinting methods can still be vulnerable to adversarial training \cite{madry2018towards} or transfer learning \cite{weiss2016survey} and are resource-intensive as well as time-consuming for the model owner \cite{lukas2020deep}.
Furthermore, when it comes to the threat of model stealing attacks on well-trained deep face recognition models, unfortunately, it is regrettable to note that no model fingerprinting methods have been proposed so far.
Faced with these stealing threats within the field of deep face recognition, we propose a model fingerprinting method tailored specifically to this domain.

To overcome the weaknesses of existing methods and solve the model fingerprinting problem in deep face recognition, we propose a correlation-based model fingerprinting method called SAC.
As mentioned above, existing model fingerprinting methods rely on the suspect model's output as a point-wise indicator to detect the stolen models, which neglects the information hidden behind pair-wise correlation.
Intuitively, samples with similar outputs in the source model are more likely to have similar outputs in the stolen models \cite{guan2022you}.
Specifically, we utilize the correlation difference between the source model and the suspect model as an indicator for detecting the stolen model.
Nevertheless, calculating correlation among clean samples from the defender's dataset may be affected by the common knowledge shared by most models trained for the same task, on which most of the models produce identical labels.
To get rid of the influence of common knowledge shared by most models, we leverage data augmentation to magnify the difference between models and have studied the influence of different augmentation methods used in \citet{hendrycks2018benchmarking} on SAC.
Results demonstrate that SAC with JPEG Compression (SAC-JC) achieves the best results on different tasks and model architectures.
Furthermore, on the task of face verification, because the model owner can only know whether two images are from the same identity and cannot get access to the exact label or probability of the images, we propose Feature from Reference Images (FRI).
FRI gathers a batch of $n+1$ images from the same identity and chooses one sample as the target for augmentation and the other samples as reference, and we leverage the results of the face verification model (whether 0 or 1) to get an  $n$ dimension vector to replace the model outputs used in SAC.
To assess the effectiveness of SAC on face recognition, we conduct experiments involving five distinct types of attacks: fine-tuning, pruning, transfer learning, model extraction, and adversarial training on two common face recognition tasks face verification \cite{taigman2014deepface} and face emotion recognition \cite{goeleven2008karolinska} across different model architectures.
Furthermore, we extend our evaluation of SAC-JC to object recognition datasets including Tiny-ImageNet and CIFAR10, also demonstrating that SAC-JC outperforms the previous methods.

Our main contributions are summarized as follows:
\begin{itemize}
\item We introduce sample correlation into model IP protection, and propose utilizing the correlation difference as a robust indicator to identify model stealing attacks, which provides a new insight into model IP protection. 
\item We disclose the model stealing risk in deep face recognition and are the first to propose the model fingerprinting method on deep face recognition tasks using SAC-JC with feature generation method FRI.
\item We study different augmented methods' influence on SAC and propose SAC-JC which leverages JPEG compression to augment data and magnify the difference between models.
\item Comprehensive results substantiate that SAC-JC not only successfully detects different model stealing attacks in face recognition tasks but also succeeds in object classification tasks, across different model architectures and datasets.
Furthermore, without training surrogate models, SAC-JC significantly reduces the computational burden, being around 34393 times faster than CAE.
\end{itemize}

Compared to the preliminary conference version \cite{guan2022you}, this manuscript has made significant improvements and extensions. The main differences can be summarized into five aspects:
1) We disclose the vulnerability of deep face recognition models to model stealing attacks and design a feature generation method FRI to fingerprint the face verification task in Section \ref{sec:FRI}.
2) We study the influence of augmented samples on SAC and propose JPEG compression based SAC-JC in Section \ref{sec:find_sample}, which is more effective and converted.
3) We extend SAC-JC to two more face-related tasks including the face verification task and face emotion recognition task KDEF, and evaluate SAC-JC on two more model stealing attacks including adversarial training and knowledge distillation in Section \ref{sec:dfr}.
4) We provide additional experimental results and in-depth analysis in Section \ref{sec:experiments}.
5) We add analysis and related work about fingerprinting on deep face recognition in Section \ref{sec:intro} and Section \ref{sec:realted_fingerprinting_face}.

\begin{figure*}
\centering
\includegraphics[width=1\textwidth]{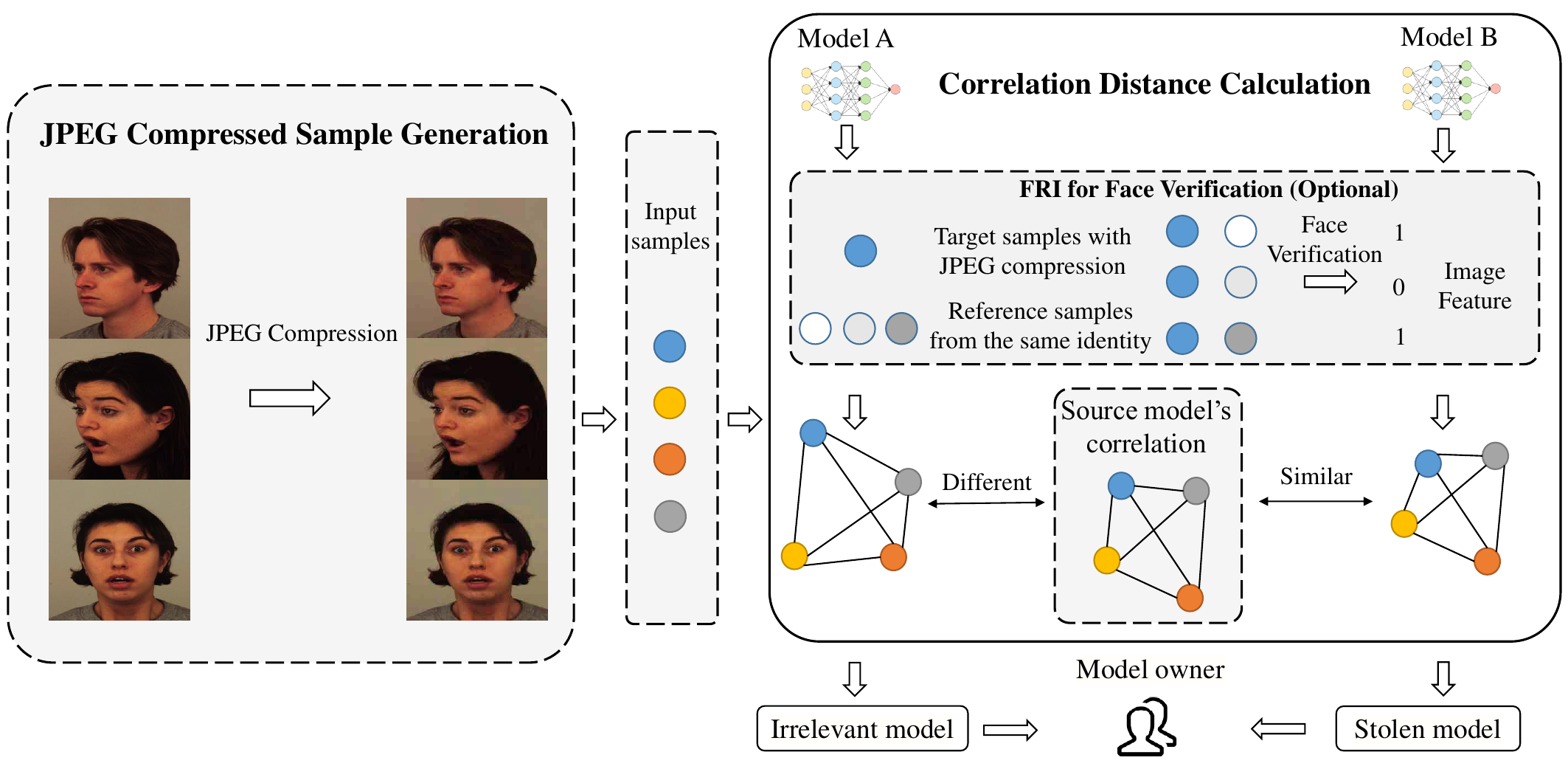}
\centering
\caption{Framework of SAC-JC. We first generate JPEG compressed samples as model inputs, represented by the colored balls. 
For the face verification model, we leverage FRI to calculate the inputs' model-specific features from the reference-target pairs.
Then we calculate the correlation difference and any suspect model with a similar correlation will be recognized as a stolen model.}
\label{fig:correlation_fingerprinting}
\end{figure*}
\section{Related Work}
% generic semantic segmentation
% Real-time semantic segmentation 
% panoptic segmentation
% light weight backbone design

\subsection{Deep IP Protection}
Model stealing attacks present a significant risk to the proprietary rights of the model's owner.
These attacks can be categorized into several distinct methods:
1. Fine-tuning \cite{tajbakhsh2016convolutional}: In this method, the attacker modifies the parameters of the source model with labeled training data for multiple epochs. The attackers can fine-tune the source model in all layers or only the last layer.
2. Pruning \cite{liu2018fine,molchanov2019importance,guan2022few}: Attackers employing this technique selectively prune less significant weights in the source model based on certain indicators, often involving activation values.
3. Transfer learning \cite{weiss2016survey}: In this setting, the attacker adapts the source model for similar tasks and utilizes the knowledge embedded in the source model to advance their own goals.
4. Model extraction \cite{orekondy2019knockoff,jagielski2020high}: Given the substantial expenses and time required for data labeling, attackers opt for this technique.
It involves replicating the functionality of the source model using unlabeled data from the same distribution.
Remarkably, this attack can be executed without access to the source model's internal parameters, relying solely on the model's outputs.
5. Adversarial training \cite{madry2018towards}: Attackers employ a blend of normal and adversarial examples to train models, which helps circumvent most fingerprinting detection methods.
To counter the threat of model stealing attacks, numerous methods for protecting model intellectual property (IP) have been proposed. These methods can generally be categorized into two main categories: watermarking methods and fingerprinting methods.

\paragraph{Watermarking Methods} Watermarking methods mainly focus on the training phase of the source model. 
They usually rely on weight regularization \cite{uchida2017embedding,chen2018deepmarks,fan2019rethinking,zhang2020passport} to add weight-related watermark into models, or train models on triggered set to leave the backdoor in them \cite{adi2018turning,zhang2018protecting}. 
Nevertheless, methods mentioned above can not detect newer attacks, for example, model extraction which trains a surrogate model from scratch \cite{orekondy2019knockoff,lukas2020deep}. Even though some watermark methods such as VEF \cite{li2022defending} or EWE \cite{jia2021entangled} could handle model extraction, they unavoidably interfere in the training process, sacrificing model utility \cite{fan2021deepip,jia2021entangled,ge2021anti} for IP protection. For VEF, it also requires white-box access to the suspect model, which greatly limits the functionality, let alone some special circumstances under which the accuracy drop for IP protection is absolutely unacceptable.

\paragraph{Fingerprinting Methods}
Fingerprinting, in contrast, capitalizes on the transferability of adversarial examples and therefore authenticates model ownership without the need to manipulate the model training process. 
This approach ensures zero compromise in model accuracy.
\citet{lukas2020deep} proposes conferrable adversarial examples, with the aim of optimizing their transferability to stolen models while minimizing the transferability to irrelevant models trained independently. 
Additionally, ModelDiff \cite{li2021modeldiff}, FUAP \cite{peng2022fingerprinting}, and DFA \cite{wang2021fingerprinting} employ various types of adversarial examples, such as DeepFool \cite{moosavi2016deepfool} and UAP \cite{moosavi2017universal}, to fingerprint the source model.
Except for this, DeepJudge \cite{chen2022copy} introduces a unified framework, utilizing various indicators to detect model stealing under both white-box and black-box settings. 
However, these techniques are vulnerable to adversarial defenses like adversarial training \cite{madry2018towards} or transfer learning, making them less effective in preserving model ownership.
Moreover, these methods often necessitate the training of numerous surrogate and irrelevant models with diverse architectures to create robust fingerprints, which imposes a significant computational burden on the model owner. 
Unlike previous approaches, our method relies on the correlation between samples rather than simple instance-level differences, enabling faster and more robust model fingerprinting with data-augmented samples instead of adversarial examples.

% Furthermore, different from Teacher Model Fingerprinting (TMF) \cite{chen2021teacher}, which relies on paired samples generated from representation-layer matching to detect transfer learning attacks, our method leverages the correlation among independent samples and therefore is able to detect a broader range of model stealing scenarios.

\subsection{Model Fingerprinting on Deep Face Recognition}\label{sec:realted_fingerprinting_face}
Deep face recognition has emerged as a prominent biometric technique for identity authentication and has been widely used in many areas, such as finance, public security, and daily life \cite{wang2021deep}.
Well-trained deep face recognition models hold significant value and can be deployed as either cloud-based services or client-side software.
However, these well-trained deep face recognition models are threatened by the model stealing attacks \cite{yu2020cloudleak} and an example is that, only with the black-box access to the source model, model extraction attacks can reach 95.0$\%$ of the original accuracy of the source model on KDEF \cite{goeleven2008karolinska} in face emotion recognition. 
Furthermore, model stealing attacks also threaten the face verification models.
Face verification is an important task in face recognition, and computes the one-to-one similarity between faces and determines whether two faces belong to the same identity \cite{lu2015surpassing}.
Because the defenders can only get access to whether two faces belong to the same identity in the suspect model while no labels are available in the prevailing black-box fingerprinting setting, there are problems with existing model IP protection methods.
% Furthermore, when it comes to face verification, in the prevailing black-box setting, the defenders can only get access to whether two faces belong to the same identity or not in the suspect model, and no labels are available, causing problems to existing model IP protection methods.
As far as we know, no existing model IP protection methods are designed for protecting the model's intellectual property in verification tasks such as face verification.  
To solve the problem in face verification, we leverage FRI to generate samples' specific features from the 0-1 results of face verification models and calculate the correlation difference between the source model and the suspect models to detect the model stealing attacks.

%%%%%%%%%%%%%%%%%%%
% \begin{figure*}[!t]
% 	\centering
% 	\includegraphics[width=1.0\linewidth]{figs/eccv_sf_up.pdf}
% 	\caption{Visualization of feature maps and semantic flow field in FAM. Feature maps are visualized by averaging along the channel dimension. Larger values are denoted by hot colors and vice versa. We use the color code proposed in~\citeppp{flowvis} to visualize the Semantic Flow field. The orientation and magnitude of flow vectors are represented by hue and saturation, respectively. As shown in this figure, using our proposed semantic flow results in more structural feature representation. }
% 	\label{fig:issue}
% \end{figure*}
%%%%%%%%%%%%%%%%%%%

\section{Method}

\subsection{Problem Definition}
% \paragraph{Attacker}
% \paragraph{Defender}
In the context of the model IP protection scenario, there are two primary parties involved: the defender and the attacker. 
The defender is the entity that owns the well-performing machine learning model by using a proprietary training dataset and a specific training algorithm \cite{cao2021ipguard}. 
In a cloud service setting, the defender deploys their well-trained models as a cloud service or client-sided software \cite{cao2021ipguard} so that attackers could only get access to the model output. Another setting is called the client-sided software setting, where the attacker can get white-box access to all the inner parameters as well as the model structure. 

The attacker's objective is to use their own dataset, which follows the same distribution as the defender's data, to reverse engineer a model that closely mimics the accuracy of the original model, while the defender aims at identifying the ownership of the suspect model. Generally speaking, cloud service deployment is the prevailing choice in the market. As such, our primary focus is on black-box IP protection, where the defender is constrained to accessing only the output of the suspect model while remaining oblivious to the architecture of the suspect model.

\subsection{Sample Correlation for Neural Network Fingerprinting}
Previous fingerprinting methods have traditionally focused solely on point-wise consistency between the source model and the suspect model, identifying model stealing behavior by determining whether the suspect model classifies the same adversarial examples into the same incorrect classes as the source model. However, these methods often prove to be less robust when it comes to scenarios involving adversarial training or transfer learning.

To enhance the robustness of model stealing detection, we shift our approach away from point-wise criteria to the pairwise relationship between model outputs.
The intuition behind this is easy to understand: when two samples yield similar outputs in the source model, they are more likely to produce similar outputs in the stolen models as well. Therefore, we introduce SAC, a correlation-based fingerprinting method that leverages the previously mentioned correlation consistency to identify model stealing effectively.
Additionally, to mitigate the impact of shared common knowledge among irrelevant models, we investigate the process of searching for suitable samples, as elaborated in Section \ref{sec:find_sample}.
An overview of our framework is illustrated in Figure \ref{fig:correlation_fingerprinting}.

Here we elaborate on the details of model fingerprinting with sample correlation \cite{ma2007discriminant,peng2019correlation}. 
Assign $O_{source} = \{ o_{1}^{source},o_{2}^{source}, \cdots, o_{n}^{source}\}$ and $O_{stolen} = \{ o_{1}^{stolen},o_{2}^{stolen}, \cdots, o_{n}^{stolen}\}$ as the set of the outputs for the source and stolen models, where $o_{k}^{source}$ and $o_{k}^{stolen}$ are the output of the $k$-$th$ input sample from the source model and stolen model, we could calculate the correlation matrix among all $n$ input samples so as to get a model-specific correlation matrix $C$ as follows: 
\begin{equation}
\begin{aligned}
    C &= \Phi (O), \quad C\in R^{n\times n} \\ where \quad C_{i,j} &= corr(o_{i},o_{j})  \quad i,j=1,\cdots,n,
\end{aligned}
\end{equation}
where $C_{i,j}$ denotes the $i,j$ entry of the model's correlation matrix $C$, which is computed by assessing the correlation between the $i$-$th$ and $j$-$th$ outputs of the model. $\Phi$ is the function for calculating the correlation matrix based on the output set.
To precisely measure the correlation among the model's outputs, we introduce several functions to model the relationship of outputs \cite{peng2019correlation}. First is cosine similarity \cite{nguyen2010cosine}, which denotes the cosine of the angle between two vectors:
\begin{equation}
\begin{aligned}
     C_{i,j}= Cos(o_{i},o_{j}) = \frac{o_{i}^{T}o_{j}}{||o_{i}||||o_{j}||},\quad i,j=1,\cdots,n.
\end{aligned}
\end{equation}
Another method for measuring the correlation of model outputs is Gaussian RBF \cite{steinwart2006explicit}. 
Gaussian RBF is a popular kernel function, computing the distance between two instances based on their Euclidean distances:
\begin{equation}
\begin{aligned}
    C_{i,j}= RBF(o_{i},o_{j}) = exp(-\frac{\Vert o_{i}-o_{j}\Vert_{2}^{2}}{2\delta^{2}}),\\ \quad i,j=1,\cdots,n.
\end{aligned}
\end{equation}
Once the correlation matrices have been computed for both the source model and the suspect model, we proceed to calculate the $L1$ distance between these matrices, which serves as our fingerprinting indicator. Any model whose distance to the source model falls below a specified threshold value, denoted as $d$, will be identified as a stolen model:
\begin{equation}
   Distance =\frac{{\Vert C_{stolen} - C_{source}\Vert}_{1}}{n^{2}} \leq d,
\end{equation}
where we denote $C_{stolen}$ and $C_{source}$ as the correlation matrix of the stolen model and the source model.
In scenarios where defenders seek to determine the optimal threshold value $d$, the use of a validation set can be instrumental. For instance, defenders can utilize the average of the means of the correlation indicators obtained from the irrelevant models and the models created through adversarial extraction on the validation set as the threshold $d$. 
% This approach helps in selecting an effective threshold for distinguishing stolen models.
% Besides, in a more restrictive scenario where the defender could only access the predicted label of the suspect model instead of detailed output probability, We propose to turn to the label-smooth technique. Smoothed label, as a soft form of the predicted label, has a similar form to the teacher's output and could therefore replace the teacher's output in knowledge distillation \cite{yuan2020revisiting}. We generate the smooth-label probability as below:
% \begin{equation}
%    p(x) = (1-\epsilon)\delta(x)+\epsilon u(x), \quad\epsilon\in[0,1],
% \end{equation}
% where $p(x)$ represents the smooth-label probability of model input $x$, $\delta(x)=[0,0,\cdots,1,\cdots,0]$ denotes the one-hot encoding for the predicted label, and u(x) is a uniform distribution.Similar to model's output probability, the smooth label proves to be effective in our model fingerprinting method SAC. It contributes to the robustness and reliability of our approach in detecting stolen models, even in scenarios with limited access to information.

\subsection{How to Find Suitable Samples?}\label{sec:find_sample}

Utilizing the correlation matrix stated above, SAC calculates the distance between the source models and the suspect models. 
In addition, it is crucial to have suitable samples as model inputs to support this fingerprinting process.
Because models trained for the same task will output the same ground-truth labels on most of the clean samples, SAC's performance is affected by the common knowledge shared by these models.

\begin{figure}
\centering
\includegraphics[width=0.47\textwidth]{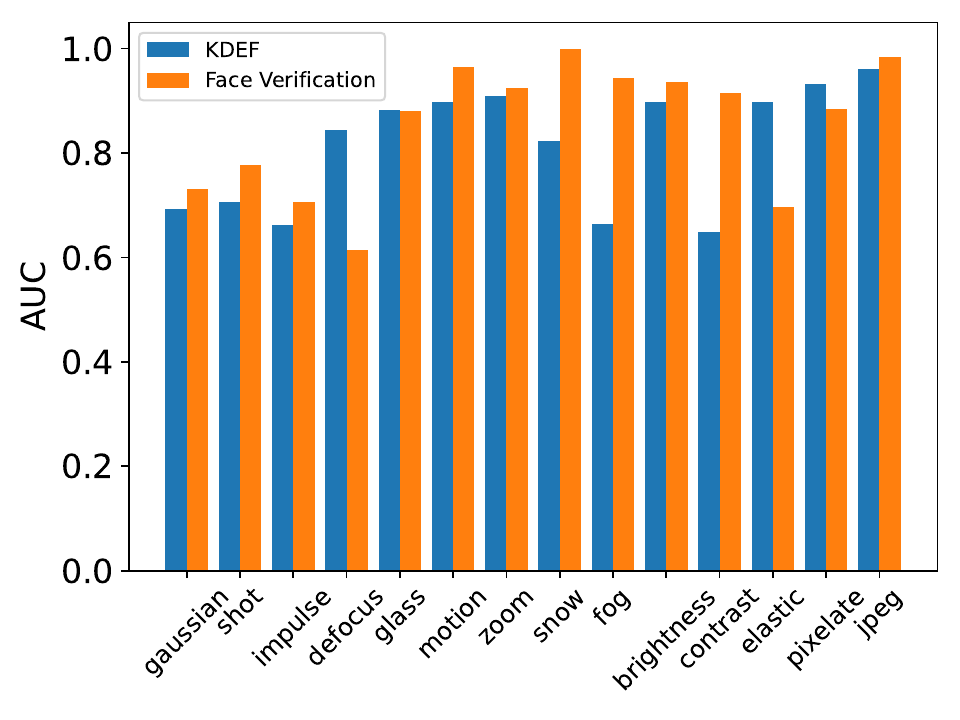}
\centering
\caption{Different image corruption methods for SAC. The results are demonstrated in terms of the average AUC across different kinds of model stealing attacks.}
\label{fig:augmentation}
\end{figure}

\paragraph{Fingerprinting with JPEG Compressed Samples}
To get rid of the influence of common knowledge and amplify the difference between models, we leverage data augmentation on the randomly selected normal samples as the input for SAC.
We first study the influence of 14 different types of corrupted methods from \citet{hendrycks2018benchmarking} as the augmented methods on SAC.
Figure \ref{fig:augmentation} demonstrates the result of SAC with different augmented methods in terms of the average AUC on different kinds of model stealing attacks.
Through the experiments, JPEG corruption (JPEG compression) has the best detection result.
JPEG corruption is a common data corruption method and it compresses a clean image in the JPEG format using different quality levels \cite{zoran2009scale}.
It includes operations including color mode conversion, downsampling, discrete cosine transform, and quantization to lower the size of images.
Different from adversarial noise, which is maliciously crafted by the attackers and is easily detected by the model provider using adversarial detection \cite{metzen2016detecting}, JPEG compression is a daily-used image compression method, and therefore, cannot be detected by the model provider.

Another advantage of leveraging JPEG compression is that JPEG corruption is not related to the models' adversarial robustness and can detect the stolen models accurately after adversarial training or adversarial extraction. 
As far as we know, all existing model fingerprinting methods rely on transferable adversarial examples to identify stolen models.
Because the success rate of adversarial examples is related to model robustness, attackers can utilize adversarial training \cite{bai2021recent} to evade these fingerprinting methods' detection.
Our experiments demonstrate that attackers can successfully evade detection by fine-tuning the extracted model for just a few epochs with unlabeled data and the predicted label from the source model in an adversarial training way, expressed as follows:
\begin{equation}
   \min \limits_{\theta_{stolen}}\sum\limits_{i} \max \limits_{||\delta||\le \epsilon}l(f_{stolen}(x+\delta),f_{source}(x)),
   \label{eq:adv_train}
\end{equation}
where $f_{stolen}$ and $f_{source}$ denotes the stolen and the source model, $\delta$ denotes the adversarial noise smaller than the bound $\epsilon$, which is generated by the attacker using adversarial attack methods such as FGSM \cite{kurakin2016adversarial} or PGD \cite{madry2018towards} and $\theta_{stolen}$ denotes the parameters of the stolen model.

\subsection{Calculating Image Feature in Face Verification}\label{sec:FRI}

In the context of face verification tasks, the model does not produce output labels. 
Instead, it only provides a binary result, typically denoting whether two images belong to the same identity.
The lack of model output in label space from images causes problems in calculating the correlation matrix in SAC.
To handle this issue as well as get the feature of the target image, we propose a method called Feature from Reference Images (FRI).
Aiming at forming the specific feature of the target image, FRI first gathers $n$ reference images with the same identity as the target image, and then, makes use of JPEG compression to augment the target image to amplify the difference between the target image and the reference images.
Finally, FRI forms $n$ target-reference pairs as the input of the suspect models and gets an $n$ dimension vector as the model-specific feature of the target image:
\begin{equation}
F_{t} = [V(I_{t},I_{r(1)}), V(I_{t},I_{r(2)}) \cdots   V(I_{t},I_{r(n)})]
\end{equation}
where $F_{t}$ represents the feature of the target image, $V$ represents the verification model, $I_{t}$ represents the target image, and $I_{r(n)}$ represents the $n-th$ reference image.
To be specific, we set $n=50$ in our experiments and $F_{t}$ is a 50 dimension 0-1 feature for the target image.
We then replace the output of the model in SAC with the feature $F_{t}$ generated from FRI.
The algorithm for employing SAC-JC with the feature generation method FRI in face verification is illustrated in Algorithm \ref{alg:SAC-JC}
The experiment results illustrated in Section \ref{sec:dfr} demonstrate the effectiveness of SAC-JC with FRI in face verification.

% \begin{algorithm}[t]
% \caption{{Pseudocode of TALL in a PyTorch-like style.}}
% \label{alg:code}

% % \algcomment{\fontsize{9pt}{0em}\selectfont \texttt{rr}: rearrange
% % }
% \definecolor{codeblue}{rgb}{0.25,0.5,0.5}
% \lstset{
%   backgroundcolor=\color{white},
%   basicstyle=\fontsize{10pt}{10pt}\ttfamily\selectfont,
%   columns=fullflexible,
%   breaklines=true,
%   captionpos=b,
%   commentstyle=\fontsize{9pt}{9pt}\color{codeblue},
%   keywordstyle=\fontsize{9pt}{9pt},
% %  frame=tb,
% }
% \begin{lstlisting}[language=python]
% # x: one clip of video (N*t*C*H*W)
% # N: number of clips
% # t: frame number of each clip
% # C: channels; s: mask size
% # x_tall: thumbnail image

% #TALL's augmentation strategy
% h = np.random.randint(H)
% w = np.random.randint(W)
% #the mask position is fixed for each frame
% m = np.ones((H, W))
% h1 = np.clip(h - s // 2, 0, H)
% h2 = np.clip(h + s // 2, 0, H)
% w1 = np.clip(w - s// 2, 0, W)
% w2 = np.clip(w + s // 2, 0, W)
% m[h1: h2, w1: w2] = 0
% m = torch.from_numpy(m)
% m = mask.expand_as(x)
% x = x * m
% #TALL: generation of the thumbnail
% x = x.view(N,-1,H,W)
% x = x.reshape(N,-1, H/sqrt(t),W/sqrt(t))
% x_tall = rearrange(x, `b (th tw c) h w -> b c (th h) (tw w)`, th=sqrt{t}, c=C)
% \end{lstlisting}
% \end{algorithm}

\begin{algorithm}[tb]
\caption{SAC-JC with FRI for fingerprinting face verification model}
\label{alg:SAC-JC}
\textbf{Input}: Source model $V_{source}$, Suspect model $V_{suspect}$, Target images $I_{t}$, Reference images $I_{r}$, Detection threshold $d$, Numbers of target images $num$ \\
\textbf{Output}: Detection result for model stealing attack $S$
\begin{algorithmic}[1] %[1] enables line numbers
\FOR{$i$ in ${1\ldots num}$}
\STATE $F_{source}^{i} = [V_{source}(I_{t}^{i},I_{r(1)}^{i}) \cdots   V_{source}(I_{t}^{i},I_{r(n)}^{i})]$
\STATE $F_{suspect}^{i} = [V_{suspect}(I_{t}^{i},I_{r(1)}^{i}) \cdots   V_{suspect}(I_{t}^{i},I_{r(n)}^{i})]$
\ENDFOR
% \STATE Calculating features of the target images on the source model:
% \FOR{$i$ in ${1\ldots num}$}
% \STATE $F_{source}^{i} = [V_{source}(I_{t},I_{r(1)}) \cdots   V_{source}(I_{t},I_{r(n)})]$
% \ENDFOR
\FOR{$m$ in ${1\ldots num}$}
\FOR{$n$ in ${1\ldots num}$}
\STATE  Calculate correlation matrix for the source model: $C_{source}(m,n) = Cos(F_{source}^{m},F_{source}^{n}) $ 
\STATE  Calculate correlation matrix for the suspect model: $C_{suspect}(m,n) = Cos(F_{suspect}^{m},F_{suspect}^{n}) $
\ENDFOR
\ENDFOR
\STATE Calculate distance: $D =\frac{{\Vert C_{suspect} - C_{source}\Vert}_{1}}{{num}^{2}}$ 
\IF {$D \leq d$}
\STATE $S=Stolen$
\ELSE
\STATE $S=Irrelevant$
\ENDIF
\end{algorithmic}
\end{algorithm}

\section{Experiment} \label{sec:experiments}
% logic
% experiment settings:
% main results 
% ablation 
%  the original SFnet ablation (only one table)
%  comparison with sfnet v2 . ablation 
% improvement analysis

\subsection{Setup}\label{sec:set_up}
In this section, we evaluate various methods for safeguarding model intellectual property (IP) against diverse model stealing attacks on a range of datasets and model structures, confirming the efficacy of SAC-JC.
To be specific, we design our evaluation of different IP protection methods against five categories of stealing attacks as listed below:
\begin{itemize}
\item \textbf{Fine-tuning.}
Typically, there exist two prevalent fine-tuning techniques: fine-tuning the last layer (Finetune-L) or fine-tuning all the layers (Finetune-A). As the names stated, Finetune-L indicates keeping the majority of the model's layers frozen and only training the final layers, while Finetune-A means fine-tuning the entire model, including all of its layers. In our experiments, we assume that the attacker fine-tunes the source model with an SGD optimizer on the attacker's dataset.
\item \textbf{Pruning.}
In our settings, we adopt Fine Pruning \cite{liu2018fine} as the pruning method.
Fine Pruning, as a commonly used method that involves pruning neurons based on their activation values, removes less significant neurons from a neural network.
As a common backdoor defense method, it could typically remove neurons that contribute to backdoor or malicious behaviors. 
Here it could serve as an attack to threat IP protection.
\item \textbf{Model Extraction.}
In the realm of model extraction attacks, there are generally two primary categories: probability-based model extraction and label-based model extraction.
Label-based model extraction attacks \cite{jagielski2020high, orekondy2019knockoff} focus on exploiting the defender's predicted labels to steal knowledge from the source model. The loss function could be expressed as $L=CE(f_{stolen}(x),l_{source})$, where $l_{source}$ indicates predicted labels from the source model and $CE(\cdot)$ is the cross-entropy loss. As for the probability-based model extraction \cite{truong2021data, jagielski2020high,gou2021knowledge}, the attacker possesses detailed output probability to train their stolen model:
\begin{equation}
\begin{aligned}
L=\alpha \cdot KL({f_{stolen}^{T}(x)},
{f_{source}^{T}(x)}) \\+(1-\alpha)\cdot CE(f_{stolen}(x),l_{source}),
\end{aligned}
\end{equation}

where $f_{stolen}^{T}(x)$ and $f_{source}^{T}(x)$ are the soft probability from the stolen and source models: $f^{T}(x) = softmax(\frac{f(x)}{T})$, in which $T$ indicates the temperature, and $KL(\cdot)$ refers to the KL divergence. In the following experiments, we fix $T=20$ as the temperature.

In face verification, because no label is available, we leverage the white-box knowledge distillation to replace the model extraction in our experiments.
Knowledge distillation \cite{gou2021knowledge} is one of the model compression methods, which utilizes the knowledge from teachers to train the student models.

\item \textbf{Adversarial Model Extraction.}
Sharing the similar logic as adaptive model extraction from CAE \cite{lukas2020deep}, the attacker could evade fingerprint detection with adversarial training after the label-based model extraction. 
The slight difference between adaptive model extraction and this thread is that the attacker achieves the extraction by the predicted label from the source model in Equation \ref{eq:adv_train}, rather than the ground-truth label.
Therefore, after adversarial training, the attacker could evade adversarial-example-based fingerprinting methods with negligible accuracy sacrifice. 
Additionally, in face verification, we leverage adversarial training as one of the attackers' methods to evade the attackers' detection.

\item \textbf{Transfer Learning.}
The attacker may also utilize the transfer learning technique to repurpose the source model for other related tasks, taking advantage of the model's knowledge while escaping from potential fingerprint detection. To simulate this, we transfer the CIFAR10 model to CIFAR10-C \cite{hendrycks2018benchmarking} and CIFAR100 dataset, from which we choose the first 10 labels. In addition, we perform transfer learning on the Tiny-ImageNet model, which is originally trained on the first 100 labels in the Tiny-ImageNet dataset, to the remaining 100 labels in the same dataset.

\end{itemize}
\paragraph{Model Architecture.}
We evaluate different IP protection methods across a range of commonly used model architectures. All extraction models and irrelevant models are trained on VGG \cite{simonyan2014very}, ResNet \cite{he2016deep}, DenseNet \cite{huang2017densely} and MobileNet \cite{sandler2018mobilenetv2} on the multi-classification tasks, including KDEF, Tiny-ImageNet, and CIFAR10.
Additionally, on the face verification task, we follow the experiment in 
a famous project Insightface~\footnote{\url{https://github.com/deepinsight/insightface}}
% \url{https://github.com/deepinsight/insightface}
and leverage the commonly used model architecture in face recognition as irrelevant models and distillation models, including ResNet18, ResNet50, and MobileFace.
% Additionally, in the context of transfer learning attacks, we employ the VGG architecture for both the transferred models and the newly-created irrelevant models. To be specific, we utilize VGG or ResNet model as the source model to perform experiments on various datasets, employing different architectures with varying accuracy levels.
Furthermore, to ensure the robustness of our results, for each attack and irrelevant model architecture, we train five models in case of randomness, in other words, 20 models for irrelevant models, extraction models, and fine-tuning models.

\paragraph{Model IP Protection Methods.} 
In order to validate the effectiveness of our method, we conduct a comparative analysis against several existing approaches, including IPGuard \cite{cao2021ipguard}, CAE \cite{lukas2020deep}, and EWE \cite{jia2021entangled}. IPGuard and CAE leverage the transferability of adversarial examples by testing the success rate of these adversarial examples when applied to the suspected models. 
If the attack success rate for any model exceeds a predefined threshold, it is identified as a stolen model.
In the face verification task, we utilize adversarial attacks designed for face recognition \cite{zhu2019generating} and calculate the attack success rate in pairs to adapt IPGuard and CAE to face verification tasks for a fair comparison. 
In contrast, EWE takes a different approach by training the source model using backdoor data \cite{gu2017badnets} and embedding a watermark within the model. 
By employing a soft nearest neighbor loss to intertwine the watermark data with the training data, EWE aims to enhance the transferability of the watermark against model extraction.
One thing to note is that we did not include the results of EWE on the face verification task and face emotion recognition task.
In face verification, there is no output label from the verification model, and thus EWE fails.
Additionally, in face emotion recognition, even if we try our best and use the official code of EWE, EWE causes the model to collapse in main tasks with accuracy dropping lower than $20\%$.

\paragraph{Datasets.} 
To assess the effectiveness and robustness of various fingerprinting methods, we perform experiments on different datasets and multiple tasks.
Following previous works, there are two datasets used in our experiments,  $D_{defender}$ and $D_{attacker}$, which belong to the defender and the attacker respectively.
In the face verification task, we leverage MS1MV2 \cite{guo2016ms} as the dataset of the defender, and CASIA-Webface \cite{yi2014learning} as the dataset of the attacker.
Additionally, we leverage ArcFace \cite{deng2019arcface} as our model training protocol.
As for the multi-classification tasks, e.g. KDEF \cite{goeleven2008karolinska}, Tiny-ImageNet \cite{le2015tiny}, and CIFAR10 \cite{krizhevsky2009learning}, according to previous methods, we split the training dataset into two equal-sized subsets: $D_{defender}$ and $D_{attacker}$, which are owned by the attacker and defender, respectively.
One point worth noting is that given the limitation of only 250 samples per label in Tiny-ImageNet, which leads to a source model accuracy drop to approximately 40\%, we opt to curate a smaller dataset by selecting the initial 100 labels. This choice allows for a higher source model accuracy.

\paragraph{Evaluation Metrics.}
To assess the effectiveness of different fingerprinting methods, similar to CAE \cite{lukas2020deep}, we employ the AUC-ROC curve \cite{davis2006relationship} and calculate the AUC value, which quantifies the separation between the fingerprinting scores of the irrelevant models and the stolen models, serving as a measure of fingerprinting effectiveness. The ROC curve is a graphical representation of the True Positive Rate and False Positive Rate. 
AUC, the area under the ROC curve, ranges from 0 to 1, with a higher AUC indicating a superior fingerprinting method. 
% When $AUC=0.5$, the fingerprinting detection can be considered as random guessing.
For further evaluating the performance of different fingerprinting methods, we follow \cite{li2022defending,peng2022fingerprinting} and introduce p-value as another evaluation metric.
We leverage an independent two-sample T-test to calculate the p-value with the null hypothesis $H_{0}:\mu_{suspect}=\mu_{irrelevant}$, where $\mu_{suspect}$ and $\mu_{irrelevant}$ represent the average of the fingerprinting scores of the suspect and irrelevant models.
To be specific, the fingerprinting score represents the correlation distance in SAC-JC, and attack success rate in IPGuard, CAE, and EWE.
A smaller p-value indicates a higher level of confidence and a better distinction between the suspect models and the irrelevant models.

AUC and p-value are both metrics not related to the threshold.
To evaluate the performance of the fingerprinting methods with a specific threshold, we introduce the F1 score as another metric.
The F1 score, which is the harmonic mean of precision and recall, can be calculated using the formula:
$F1 = 2\times \frac{Precision \times Recall}{Precision + Recall}$.
Additionally, we have provided the details of the specific threshold selection in the following paragraph.

\paragraph{Threshold Selection.}

In general, thresholds can be determined using a small validation set.
We follow the threshold decision method used in \cite{li2021modeldiff} and select the worst value found in the irrelevant models.
To be specific, we choose the smallest correlation distance found in irrelevant models in SAC-JC and the highest attack success rate found in irrelevant models in IPGuard, CAE, and EWE as the threshold.
To reduce the need for collecting irrelevant models, we only use four irrelevant models in our experiments across different datasets and source model architectures. 
Additionally, since model fingerprinting is a black-box detection method and we do not know the category of the model stealing attacks, we use the same threshold for all suspect models within one task for detection.

% Moreover, to better illustrate the impact of model architectures and initialization on fingerprinting, we introduce the concept of a "fingerprinting score" (Score). This Score serves as an indicator that different methods use to distinguish suspect models. To elaborate, the Score represents the attack success rate in IPGuard, CAE, and EWE, while it represents the correlation difference between the suspect model and the source model in SAC-JC.

% \begin{table}[t!]\footnotesize
%  \caption{Different model IP protection methods distinguish irrelevant and stolen models on KDEF.}
%   \centering
%   \setlength{\tabcolsep}{1mm}{
%     \begin{tabular}{c c c c c c }  
%     \hline  
%    {Attack(AUC$\uparrow$)}&  
%     {IPGuard \cite{cao2021ipguard}}&{CAE \cite{lukas2020deep}} &{EWE \cite{jia2021entangled}} & {SAC-JC}  \cr
%     % \cr\cline{2-11}   \rowcolor{white}
   
%     \hline  
%     Finetune-A & $1.00$   & $1.00$ & $-$ &   $1.00$   \cr
%     Finetune-L & $1.00$   & $1.00$  & $-$  &  $1.00$ \cr
%     Pruning & $0.95$   & $0.85$  & $-$  &  $0.99$    \cr
%     Extract-L & $0.61$   & $0.77$ & $-$  &  $0.92$   \cr
%     Extract-P & $0.61$  & $0.82$  & $-$  &  $0.97$  \cr
%     Extract-Adv & $0.36$   & $0.58$  & $-$  &  $0.93$   \cr\hline
%     \textbf{Average} & $0.76$  & $0.84$  & $-$  &  $0.97$  \cr\hline
    
%     \end{tabular}  
%     }

%     \label{tab:fingerprint_kdef}  
% \end{table} 

\begin{table*}[t!]\footnotesize
 \caption{Different model IP protection methods distinguish irrelevant and stolen models on KDEF.}
  \centering
\setlength{\tabcolsep}{0.8mm}{
     \begin{tabular}{c c l c c l c c c c c l c}  
    \hline  
   {Attack}&  
    \multicolumn{3}{c} {IPGuard \cite{cao2021ipguard}}& \multicolumn{3}{c}{CAE \cite{lukas2020deep}} & \multicolumn{3}{c}{EWE \cite{jia2021entangled}} &  \multicolumn{3}{c}{SAC-JC}  
    % \cr\cline{2-11}   \rowcolor{white}

     \cr\cline{2-13}   
     & AUC $\uparrow$ & \multicolumn{1}{c}{p-value $\downarrow$} & F1 $\uparrow$ & AUC $\uparrow$ & \multicolumn{1}{c}{p-value $\downarrow$} & F1 $\uparrow$ & AUC $\uparrow$ & \multicolumn{1}{c}{p-value $\downarrow$} & F1 $\uparrow$ & AUC $\uparrow$ &\multicolumn{1}{c}{p-value $\downarrow$} & F1 $\uparrow$ 
   
   \cr \hline  
    Finetune-A & $1.00$ & $3.33*10^{-15}$ & 0.87   & $1.00$& $1.13*10^{-6}$ & 0.83  & $-$& $-$ & $-$ &   $1.00$ & $6.02*10^{-18}$ & 0.95 \cr
    Finetune-L & $1.00$ & $1.26*10^{-15}$ &  0.87 & $1.00$ & $1.94*10^{-7}$ &  0.83 &  $-$& $-$ & $-$  &  $1.00$ & $2.01*10^{-19}$& 0.95\cr
    Pruning & $0.95$ & $3.61*10^{-7}$  & 0.76  & $0.85$  & $6.88*10^{-4}$ & 0.67 & $-$  & $-$& $-$ &  $0.99$ & $2.98*10^{-5}$& 0.95   \cr
    Extract-L & $0.61$ & $2.14*10^{-1}$ &  0.36 & $0.77$  & $7.81*10^{-4}$ & 0.59 & $-$& $-$& $-$   &  $0.92$  & $1.54*10^{-7}$& 0.84 \cr
    Extract-P & $0.61$ & $2.34*10^{-1}$ & 0.36  & $0.82$ & $1.17*10^{-4}$ &0.70 & $-$ & $-$  & $-$&  $0.97$ & $1.83*10^{-9}$ & 0.90 \cr
    Extract-Adv & $0.36$ & $3.72*10^{-2}$  & 0.00  & $0.58$ & $2.85*10^{-1}$ & 0.15 &$-$ & $-$ & $-$ &  $0.93$  & $8.13*10^{-8}$ & 0.88\cr\hline
    \textbf{Average} & $0.76$ & $8.09*10^{-2}$ & 0.54  & $0.84$ & $4.77*10^{-2}$  & 0.63 & $-$ & $-$ & $-$ &  $0.97$ & $5.00*10^{-6}$ & 0.91\cr\hline
    
    \end{tabular}  
    
}
   
    \label{tab:fingerprint_kdef}  
\end{table*}

\begin{table*}[t!]\footnotesize
 \caption{Different model IP protection methods distinguish irrelevant and stolen models on face verification.}
  \centering
  \setlength{\tabcolsep}{0.8mm}{
   \begin{tabular}{c c l c c l c c c c c l c}  
        \hline  
    {Attack}&  
    \multicolumn{3}{c} {IPGuard \cite{cao2021ipguard}}& \multicolumn{3}{c}{CAE \cite{lukas2020deep}} & \multicolumn{3}{c}{EWE \cite{jia2021entangled}} &  \multicolumn{3}{c}{SAC-JC}  
    % \cr\cline{2-11}   \rowcolor{white}

    \cr\cline{2-13}   
     & AUC $\uparrow$ & \multicolumn{1}{c}{p-value $\downarrow$} & F1 $\uparrow$ & AUC $\uparrow$ & \multicolumn{1}{c}{p-value $\downarrow$} & F1 $\uparrow$ & AUC $\uparrow$ & \multicolumn{1}{c}{p-value $\downarrow$} & F1 $\uparrow$ & AUC $\uparrow$ &\multicolumn{1}{c}{p-value $\downarrow$} & F1 $\uparrow$
   
   \cr \hline

    Finetune-A & $1.00$ & $1.32*10^{-15}$ & 0.91 & $1.00$ & $3.59*10^{-9}$ & 0.95 & $-$  & $-$& $-$ &   $0.99$  &  $6.98*10^{-6}$ & 1.00\cr
    Finetune-L & $1.00$ & $4.80*10^{-23}$& 0.91   & $1.00$& $1.29*10^{-25}$  & 0.95 & $-$& $-$& $-$ &   $1.00$ & $1.70*10^{-9}$ & 1.00\cr
    Pruning & $0.95$   & $2.28*10^{-15}$& 0.86 &  $0.94$& $1.06*10^{-3}$  & 0.60 & $-$ & $-$& $-$  &  $1.00$ & $5.32*10^{-4}$ & 1.00 \cr
    KD & $0.70$& $3.57*10^{-2}$  & 0.30 & $0.99$ & $1.11*10^{-9}$ & 0.93 & $-$& $-$ & $-$ &  $0.92$   & $2.04*10^{-5}$& 0.81\cr
    Adv-Train & $0.81$ & $3.33*10^{-2}$ & 0.44  & $0.00$& $2.41*10^{-8}$ & 0.00 & $-$ & $-$ & $-$  &  $1.00$  & $2.06*10^{-4}$ & 1.00\cr\hline
    \textbf{Average} & $0.89$ & $1.38*10^{-2}$ & 0.68 & $0.79$  & $2.11*10^{-4}$  & 0.69 & $-$& $-$& $-$  &  $0.98$ & $1.53*10^{-4}$ & 0.96\cr\hline
    
    \end{tabular}  
    }
   
    \label{tab:fingerprint_fv}  
\end{table*} 

\begin{figure*}
\centering
\includegraphics[width=0.95\textwidth]{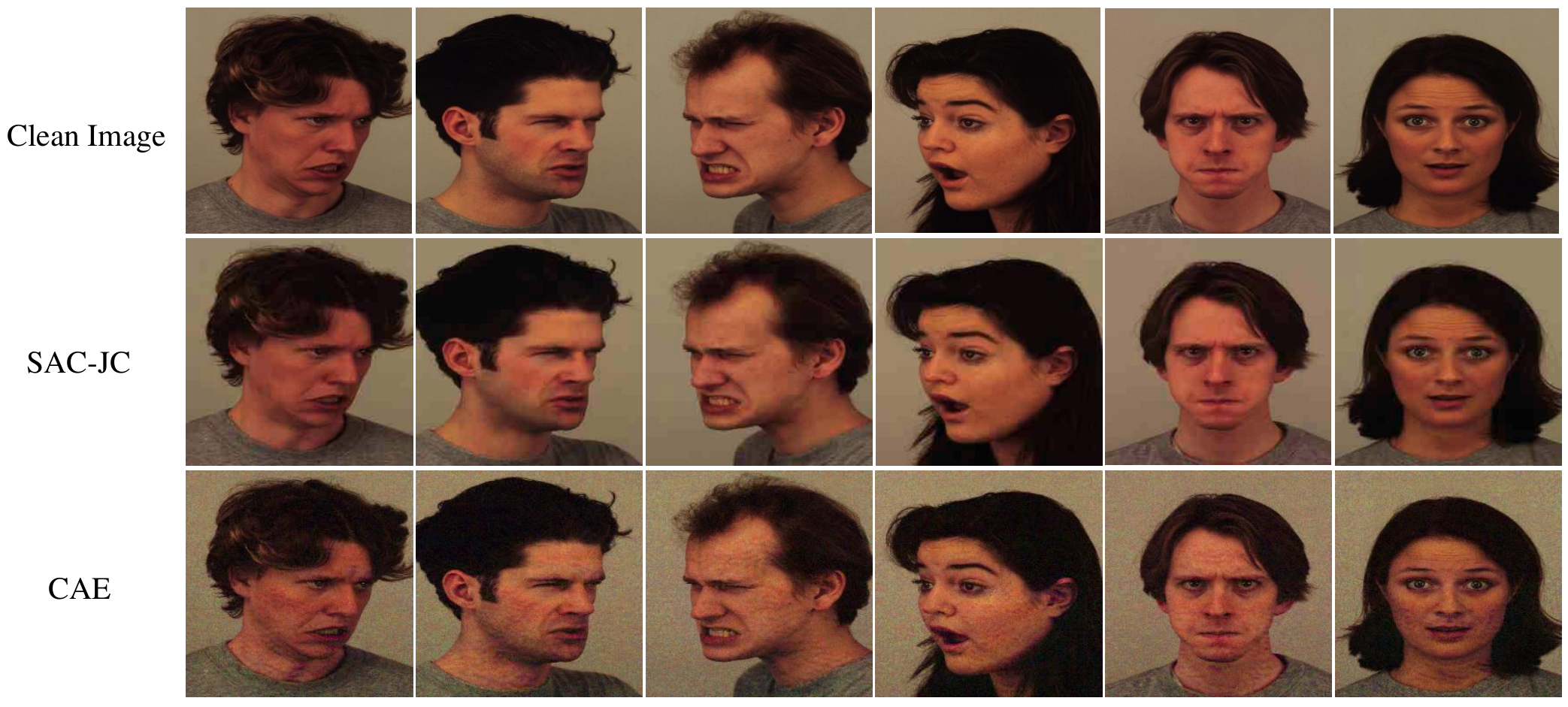}
\centering
\caption{Images used for model fingerprinting in KDEF.}
\label{fig:image_comparison}
\end{figure*}
\begin{table*}\footnotesize
 \caption{Accuracies (\%) of the source model, irrelevant models, and different attack models on KDEF.}
  \centering  
  \setlength{\tabcolsep}{0.4mm}{
    \begin{tabular}{|c| c| c| c |c |c |c |c |c| c| c| c|c|}  
    \hline  
    {$\%$}&  
    \multicolumn{4}{c|}{Irrelevant}&\multicolumn{2}{c|}{Finetune}&\multicolumn{3}{c|}{Pruning}&\multicolumn{2}{c|}{} \cr\hline  
    {Source} & VGG & ResNet & Dense & Mobile & F-A & F-L &p=0.1&p=0.2&p=0.25&\multicolumn{2}{c|}{}\cr \hline 
    
   86.19&82.67&77.06&74.27&69.30&90.41&88.14&80.96&75.58&69.19&\multicolumn{2}{c|}{}\cr\hline
   \multicolumn{4}{|c|}{Extract-L}&\multicolumn{4}{c|}{Extract-P}&\multicolumn{4}{c|}{Extract-Adv}\cr\hline
   VGG & ResNet & Dense & Mobile & VGG & ResNet & Dense & Mobile& VGG & ResNet & Dense & Mobile\cr\hline
   81.86&81.19&81.77&80.41&82.88&80.87&82.47&80.96&73.02&76.83&74.27&69.30\cr\hline

    \end{tabular}  
    }
       \label{tab:accuracy_kdef}  
\end{table*} 

\subsection{Fingerprinting on Face Recognition} \label{sec:dfr}

In this section, we evaluate SAC-JC on two common face recognition tasks: face verification and face emotion recognition.
Tables \ref{tab:fingerprint_kdef} and  \ref{tab:fingerprint_fv} demonstrate different fingerprinting methods against different model stealing attacks on face emotion recognition KDEF and face verification.
\textbf{Finetune-A} and \textbf{Finetune-L} represent fine-tuning the source model on all the layers and the last layer, and \textbf{Extract-L}, \textbf{Extract-P} and \textbf{Extract-Adv} represent the three settings of the model extraction, label-based model extraction, probability-based model extraction, and adversarial model extraction.
Besides, \textbf{KD} stands for knowledge distillation, where the source model serves as the teacher, transferring its knowledge to another model.
On the other hand, \textbf{Adv-Train} involves utilizing adversarial training with the source model.
In addition, to facilitate a more comprehensive comparison, we calculate the average AUC of different intellectual property protection methods applied to different model stealing attacks.
Experiments demonstrate the effectiveness and superior performance of our method SAC-JC in terms of AUC, p-value, and F1.
To be specific, SAC-JC achieves $AUC=0.97$ on KDEF and $AUC=0.98$ on face verification on average.
Additionally, our method SAC-JC detects model stealing attacks successfully with black-box access to both the source model and the suspect models, while the other fingerprinting methods such as IPGuard and CAE need white-box access to the source model to generate adversarial examples.
SAC-JC offers a broader range of applicability, allowing a third party to detect model stealing attacks without access to the inner parameters of the source model.

From our experiments, we observe that the average attack success rate of CAE is higher than that of IPGuard, suggesting that CAE exhibits superior transferability compared to IPGuard.
Additionally, in Table \ref{tab:fingerprint_kdef}, our findings indicate that CAE outperforms IPGuard in identifying model extraction attacks, primarily owing to the introduction of conferrable scores. 
However, it's worth noting that the success rates of these methods still exhibit significant fluctuations across various model architectures, causing the low AUC in our multi-model architecture scenario.
The success rate of attacks involving adversarial examples can be influenced by the robustness of the target model. 
Models designed with greater robustness or those that have undergone adversarial training may exhibit lower attack success rates when compared to irrelevant models.
In contrast, the correlation difference metric is independent of model robustness and excels in its ability to detect stolen models consistently across various model architectures.
Thus, our proposed method, SAC-JC, demonstrates a higher AUC compared to the other two fingerprinting methods. 
Furthermore, SAC-JC eliminates the need for the defender to train any surrogate models, which saves the defender’s time on a large scale.
We will discuss it in detail in Section \ref{sec:influence}. 

Moreover, we also compare the images used for fingerprinting in Figure \ref{fig:image_comparison}. 
Adversarial examples used in CAE or IPGuard are not only easy to attract the attention of attackers, but also easy to detect and remove by attackers \cite{lukas2020deep}.
On the contrary, SAC-JC only leverages JPEG-compressed images as the input of the suspect models and can be hardly detected or observed by the owner of the suspect model.
Additionally, in Table \ref{tab:accuracy_kdef}, we present the average accuracy of both the source and stolen models. 
This table illustrates that the majority of model stealing attacks have the capability to successfully replicate the source model with only a minimal decrease in accuracy.
To be specific, the attackers achieve $95.0\%$ of the accuracy of the source model only by utilizing the unlabeled data and the hard label from the source model.

\subsection{Fingerprinting on Object Classification}

\begin{table*}[t!]\footnotesize
 \caption{Different model IP protection methods distinguish irrelevant and stolen models on Tiny-ImageNet.}
  \centering
  \setlength{\tabcolsep}{0.6mm}{
    \begin{tabular}{c c l c c l c c l c c l c}  
    \hline  
   {Attack}&  
    \multicolumn{3}{c} {IPGuard \cite{cao2021ipguard}}& \multicolumn{3}{c}{CAE \cite{lukas2020deep}} & \multicolumn{3}{c}{EWE \cite{jia2021entangled}} &  \multicolumn{3}{c}{SAC-JC}  
    % \cr\cline{2-11}   \rowcolor{white}

     \cr\cline{2-13}   
     & AUC $\uparrow$ & \multicolumn{1}{c}{p-value $\downarrow$}  & F1 $\uparrow$ & AUC $ \uparrow$ &  \multicolumn{1}{c}{p-value $\downarrow$}  & F1 $\uparrow$ & AUC $\uparrow$ &  \multicolumn{1}{c}{p-value $\downarrow$}  & F1 $\uparrow$ & AUC $ \uparrow$ & \multicolumn{1}{c}{p-value $\downarrow$}  & F1 $\uparrow$ 
   
   \cr \hline

   Finetune-A & $1.00$ & $3.95*10^{-33}$ & 0.83  & $1.00$& $1.06*10^{-26}$ & 0.87 & $0.48$& $4.89*10^{-1}$ & 0.00 &   $1.00$ & $1.20*10^{-22}$ & 0.92 \cr
    Finetune-L & $1.00$ & $2.91*10^{-34}$ & 0.83  & $1.00$ & $1.06*10^{-26}$ & 0.87 & $1.00$ & $4.15*10^{-35}$ & 0.95 &  $1.00$& $1.17*10^{-25}$ & 0.92   \cr
    Pruning & $1.00$  & $1.33*10^{-22}$ & 0.67 & $1.00$ & $5.49*10^{-18}$ & 0.73 & $0.58$  & $2.11*10^{-1}$ & 0.32&  $1.00$ & $2.06*10^{-12}$ & 0.92  \cr
    Extract-L & $0.97$ & $4.77*10^{-10}$ & 0.86  & $1.00$& $2.49*10^{-19}$ & 0.93 & $1.00$ & $2.33*10^{-18}$ & 0.98 &  $1.00$ & $2.53*10^{-10}$ & 0.92   \cr
    Extract-P & $0.97$ & $3.56*10^{-6}$ & 0.86 & $1.00$& $3.17*10^{-14}$ & 0.93  & $1.00$ & $1.49*10^{-13}$ & 0.98 &  $1.00$ & $2.37*10^{-13}$ & 0.92 \cr
    Extract-Adv & $0.65$  & $1.23*10^{-1}$ & 0.35 & $0.78$ & $5.95*10^{-4}$ & 0.69 & $1.00$& $2.04*10^{-14}$ & 0.98  &  $0.96$& $3.65*10^{-8}$ & 0.92   \cr
    \textbf{Average} & $0.93$ & $2.04*10^{-2}$ & 0.73 & $0.96$& $9.92*10^{-5}$ & 0.84  & $0.84$ & $1.17*10^{-1}$ & 0.70 &  $0.99$  & $6.12*10^{-9}$ & 0.92\cr\hline
    Transfer-A & $-$   & \multicolumn{1}{c}{$-$}  &  \multicolumn{1}{c}{$-$}  & $-$   &  \multicolumn{1}{c}{$-$}  &  \multicolumn{1}{c}{$-$} & $-$   &  \multicolumn{1}{c}{$-$}  &  \multicolumn{1}{c}{$-$} &   $1.00$  & $1.08*10^{-3}$ & $1.00$  \cr
    Transfer-L  & $-$   & \multicolumn{1}{c}{$-$}  &  \multicolumn{1}{c}{$-$}  & $-$   &  \multicolumn{1}{c}{$-$}  &  \multicolumn{1}{c}{$-$} & $-$   &  \multicolumn{1}{c}{$-$}  &  \multicolumn{1}{c}{$-$} &   $1.00$ & $6.89*10^{-4}$ & $1.00$ \cr\hline
    
    \end{tabular}  
    }
    \label{tab:fingerprint_imagenet}  
\end{table*} 

\begin{table*}[t!]\footnotesize
 \caption{Different model IP protection methods distinguish irrelevant and stolen models on CIFAR10.}
  \centering
 \setlength{\tabcolsep}{0.6mm}{
 \begin{tabular}{c c l c c l c c l c c l c}  
    \hline  
   {Attack}&  
    \multicolumn{3}{c} {IPGuard \cite{cao2021ipguard}}& \multicolumn{3}{c}{CAE \cite{lukas2020deep}} & \multicolumn{3}{c}{EWE \cite{jia2021entangled}} &  \multicolumn{3}{c}{SAC-JC}  
    % \cr\cline{2-11}   \rowcolor{white}

     \cr\cline{2-13}   
   & AUC $\uparrow$ & \multicolumn{1}{c}{p-value $\downarrow$}  & F1 $\uparrow$ & AUC $ \uparrow$ &  \multicolumn{1}{c}{p-value $\downarrow$}  & F1 $\uparrow$ & AUC $\uparrow$ &  \multicolumn{1}{c}{p-value $\downarrow$}  & F1 $\uparrow$ & AUC $ \uparrow$ & \multicolumn{1}{c}{p-value $\downarrow$}  & F1 $\uparrow$ 
   
   \cr \hline  
  
   Finetune-A & $1.00$ & $6.50*10^{-13}$ & 0.83 & $1.00$ & $7.47*10^{-8}$ & 0.87 & $1.00$& $6.07*10^{-24}$ & 0.80 &  $1.00$& $1.85*10^{-28}$ & 1.00  \cr
    Finetune-L & $1.00$ & $4.74*10^{-13}$ & 0.83 & $1.00$ & $7.47*10^{-8}$ & 0.87 & $1.00$  & $1.33*10^{-20}$ & 0.80&  $1.00$  & $1.24*10^{-31}$ & 1.00 \cr
    Pruning & $1.00$& $1.47*10^{-11}$ & 0.83  & $0.95$ & $2.20*10^{-6}$ & 0.82 & $0.87$& $1.42*10^{-2}$ & 0.75  &  $1.00$& $1.72*10^{-19}$ & 1.00    \cr
    Extract-L & $0.81$ & $3.60*10^{-2}$ & 0.35 & $0.83$  & $8.41*10^{-5}$ & 0.36& $0.97$ & $1.65*10^{-5}$ & 0.89 &  $1.00$ & $8.43*10^{-16}$ & 1.00  \cr
    Extract-P & $0.81$ & $4.59*10^{-3}$ & 0.35  & $0.90$ & $4.95*10^{-6}$ & 0.47& $0.97$& $5.51*10^{-6}$ & 0.89 &  $1.00$ & $4.62*10^{-21}$ & 1.00\cr
    Extract-Adv & $0.54$ & $9.93*10^{-1}$ & 0.08 & $0.52$  & $8.82*10^{-1}$ & 0.08& $0.91$ & $9.65*10^{-5}$ & 0.84 &  $1.00$ & $8.14*10^{-14}$ & 0.98 \cr
    \textbf{Average} & $0.86$ & $1.72*10^{-1}$ & 0.55 & $0.87$ & $1.47*10^{-1}$ & 0.58 & $0.95$  & $2.93*10^{-3}$ & 0.83&  $1.00$ & $1.37*10^{-14}$ & 1.00\cr\hline
    Transfer-10C & $1.00$ & $3.86*10^{-4}$ & 0.95& $1.00$ & $1.41*10^{-5}$ & 0.91 & $1.00$ & $1.57*10^{-10}$ & 0.95 &  $1.00$& $6.78*10^{-12}$ & 1.00  \cr
    Transfer-A & $-$   & \multicolumn{1}{c}{$-$}  &  \multicolumn{1}{c}{$-$}  & $-$   &  \multicolumn{1}{c}{$-$}  &  \multicolumn{1}{c}{$-$} & $-$   &  \multicolumn{1}{c}{$-$}  &  \multicolumn{1}{c}{$-$} &  $1.00$  & $1.85*10^{-9}$ & 0.95 \cr
    Transfer-L   & $-$   & \multicolumn{1}{c}{$-$}  &  \multicolumn{1}{c}{$-$}  & $-$   &  \multicolumn{1}{c}{$-$}  &  \multicolumn{1}{c}{$-$} & $-$   &  \multicolumn{1}{c}{$-$}  &  \multicolumn{1}{c}{$-$}  &  $1.00$ & $6.50*10^{-12}$ & 0.95  \cr\hline
    
    \end{tabular}  
    }
    \label{tab:fingerprint_CIFAR10}  
\end{table*} 
%%%%%%%%%%%%%%%%%%%%%%%%%%%%%%%%%%%%%%%%%%%%%%%%%%%%%%%%%%%%%%%%%%%%%%%%%%%%%%%%%%%%%%%%%%%%%%%%%%%%% 
To better assess the performance of SAC-JC in object classification tasks, we conduct experiments on Tiny-ImageNet \cite{le2015tiny} and CIFAR10 \cite{krizhevsky2009learning} in Tables \ref{tab:fingerprint_imagenet} and \ref{tab:fingerprint_CIFAR10}.
In these two tables, \textbf{Finetune-A} and \textbf{Finetune-L} represent fine-tuning the source model on all the layers and the last layer respectively, while \textbf{Extract-L}, \textbf{Extract-P} and \textbf{Extract-Adv} represent the three settings of the model extraction, label-based model extraction, probability-based model extraction and adversarial model extraction.
Besides, \textbf{Transfer-A} and \textbf{Transfer-L} represents transferring the source model to a new dataset by fine-tuning all the layers or the last layer, and  \textbf{Transfer-10C} represents transferring the source model to CIFAR10C.
Similar to the results on face emotion recognition or face verification, SAC-JC performs better than IPGuard and CAE,  especially when facing model stealing attacks such as adversarial training or adversarial model extraction.
In addition, our experiments also reveal that another compared method EWE exhibits sensitivity to pruning, although its decline in AUC is less pronounced compared to other watermarking methods based on normal backdoors.
One thing to note is that all three IP protection methods, IPGuard, CAE, and EWE, cannot detect the transfer-based model stealing attacks due to the label space change in transfer learning.
These methods rely solely on the attack success rate as their identification criterion and consequently, they struggle to identify stolen models when dealing with transfer learning or label changes. 
Conversely, our methods utilize correlation differences as fingerprints, and this correlation consistency remains intact even when models are transferred to different tasks.

\subsection{Sensitivity Analysis}
To assess the efficacy of SAC-JC across varying numbers of augmented images, we conducted experiments employing SAC-JC on KDEF, Tiny-ImageNet, and CIFAR10. 
The results are depicted in Figure \ref{fig:auc_sample_num}. 
In the figures, AUC-P, AUC-L, and AUC-Adv represent the performance of SAC-JC in terms of AUC on probability-based model extraction, label-based model extraction, and adversarial model extraction respectively.
In addition, AUC-Finetune, AUC-Prune, and AUC-Transfer represent AUC of SAC-JC on fine-tuning, pruning, and transfer learning respectively.
The outcomes of this experiment highlight the robustness of our SAC method to the number of samples as model fingerprints, even in a few-shot setting. 
With just 50 or even only 25 JPEG compressed images, SAC-JC effectively detects various types of model stealing attacks with a high AUC.

\subsection{Robustness of SAC-JC on Different Model Architectures}

\begin{figure}[]
		\centering
		\small
	
			\includegraphics[width=0.85\linewidth]{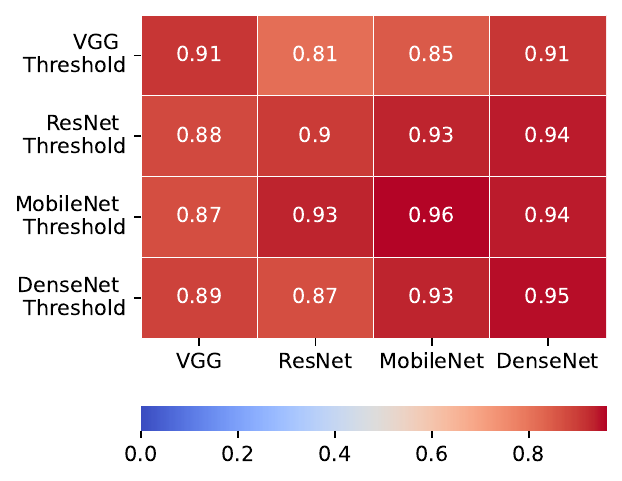} 
			
        \vspace{1pt}
		\caption{Average F1 score of SAC-JC on different source model architectures with different thresholds calculated from other source model architectures.}
		\label{fig:F1_score}
		% \vspace{-10pt}
\end{figure}

In this section, we will discuss the performance of SAC-JC on various source model architectures.
To be specific, we evaluate SAC-JC on the source model with VGG, ResNet, MobileNet, and DenseNet. 
The results are presented in Table \ref{tab:architecture}.
The experiment results have demonstrated that SAC-JC achieves good performance across different source model architectures in terms of metrics including AUC, p-value, and F1 score.
Additionally, we analyze the robustness of the selected thresholds and investigate whether a threshold determined for one model architecture can be transferred to other source model architectures.
We present the value of the threshold calculated on different source model architectures on KDEF for SAC-JC in Table \ref{tab:threshold}.
The calculated thresholds range from 0.295 to 0.321, with the correlation distance being an indicator ranging from 0 to 1.
This indicates a very small variation in the selected thresholds across different source model architectures.
Furthermore, we depict the average F1 score on different model architectures with the threshold calculated on other source model architectures in Figure \ref{fig:F1_score}.
The results further illustrate that thresholds calculated on one source model architecture can be effectively transferred to detect model stealing attacks on other source model architectures.

\begin{table*}[h]\footnotesize
 \caption{Peformance of SAC-JC on Different Model Architectures on KDEF.}
  \centering
\setlength{\tabcolsep}{0.8mm}{

     \begin{tabular}{c c l c c l c c l c c l c}  
    \hline  
   {Attack}&  
    \multicolumn{3}{c} {VGG}& \multicolumn{3}{c}{ResNet} & \multicolumn{3}{c}{MobileNet} &  \multicolumn{3}{c}{DenseNet}  
    % \cr\cline{2-11}   \rowcolor{white}

     \cr\cline{2-13}   
     & AUC $\uparrow$ & \multicolumn{1}{c}{p-value $\downarrow$} & F1 $\uparrow$ & AUC $\uparrow$ & \multicolumn{1}{c}{p-value $\downarrow$}& F1 $\uparrow$ & AUC $\uparrow$ & \multicolumn{1}{c}{p-value $\downarrow$} & F1 $\uparrow$ & AUC $\uparrow$ & \multicolumn{1}{c}{p-value $\downarrow$} & F1 $\uparrow$ 
   
   \cr \hline  
    Finetune-A &    $1.00$ & $6.02*10^{-18}$ & 0.95 &    $1.00$ & $7.67*10^{-14}$ & 0.92  &    $1.00$ & $3.95*10^{-20}$ & 0.97 & 
    $1.00$ & $1.43*10^{-16}$ & 1.00  \cr
    Finetune-L   &  $1.00$ & $3.95*10^{-20}$& 0.95&    $1.00$ & $8.47*10^{-24}$ & 0.92 &    $1.00$ & $8.51*10^{-23}$ & 0.97
    &    $1.00$ & $3.30*10^{-22}$ & 1.00\cr
    Pruning  &  $0.99$ & $2.98*10^{-5}$& 0.95 &    $1.00$ & $5.56*10^{-5}$ & 0.92  &    $1.00$ & $9.15*10^{-10}$ & 0.97
    &    $1.00$ & $2.85*10^{-6}$ & 0.95\cr
    Extract-L   &  $0.92$  & $1.54*10^{-7}$& 0.84 &    $0.98$ & $7.66*10^{-10}$ & 0.90 &    $0.97$ & $1.05*10^{-10}$ & 0.93
    &    $0.92$ & $1.24*10^{-7}$ & 0.85\cr
    Extract-P  & $0.97$ & $1.83*10^{-9}$ & 0.90 &    $1.00$ & $2.12*10^{-14}$ & 0.92 &    $0.99$ & $5.67*10^{-14}$ & 0.97
    &    $0.98$ & $4.38*10^{-12}$ & 0.95\cr
    Extract-Adv  &  $0.93$  & $5.67*10^{-14}$ & 0.88&    $0.93$ & $1.15*10^{-8}$ & 0.81 &    $0.96$ & $7.60*10^{-10}$ & 0.91
    &    $0.90$ & $4.61*10^{-7}$ & 0.89\cr\hline
    \textbf{Average}  &  $0.97$ & $5.00*10^{-6}$ & 0.91&    $0.98$ & $9.27*10^{-6}$ & 0.90 &    $0.99$ & $2.97*10^{-10}$ & 0.96
    &    $0.97$ & $5.73*10^{-7}$ & 0.95\cr\hline
    
    \end{tabular}  
    
}
   
    \label{tab:architecture}
\end{table*}

% \begin{figure*}

%     \centering
%     \subfigure[KDEF]{
% 	\includegraphics[width=0.3\linewidth]{AUC-KDEF.pdf}
% 	}
% 	\quad
%     \subfigure[Tiny-ImageNet]{
% 	\includegraphics[width=0.3\linewidth]{AUC-ImageNet.pdf}
% 	}
%  \quad
%  \subfigure[CIFAR10]{
% 	\includegraphics[width=0.3\linewidth]{AUC-CIFAR10.pdf}
	
% 	}
%     \caption{Performance change of SAC-JC with different data amounts.}
%     \label{fig:auc_sample_num}
% \end{figure*}

\begin{figure*}[h]
		\centering
		\small
		\setlength\tabcolsep{1mm}
		\renewcommand\arraystretch{0.1}
		\begin{tabular}{cccc}
			\includegraphics[width=0.32\linewidth]{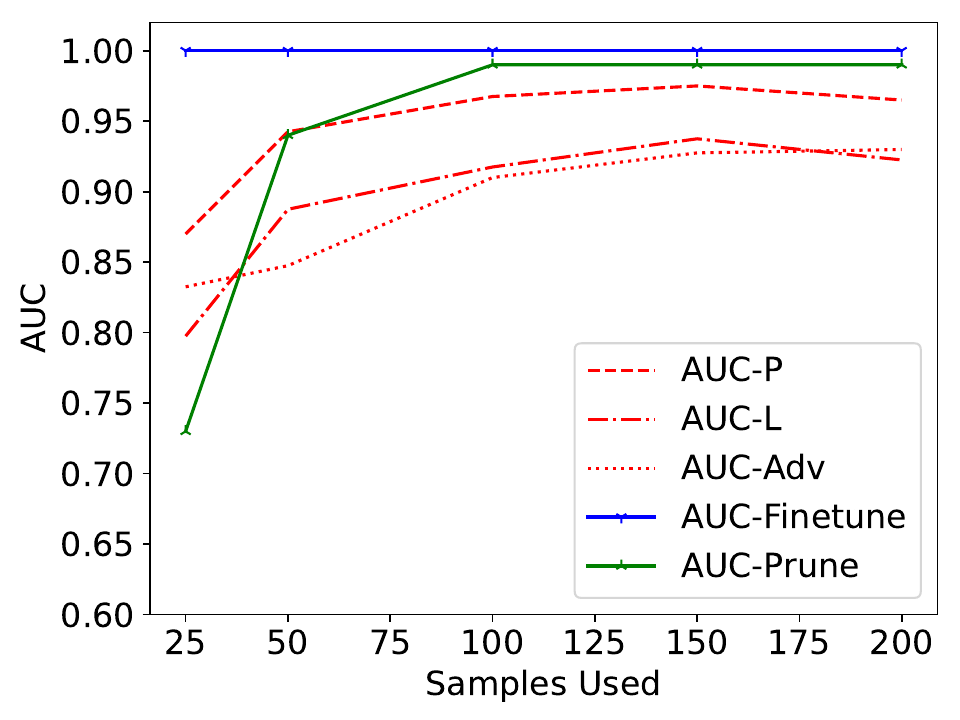} &
			\includegraphics[width=0.32\linewidth, clip]{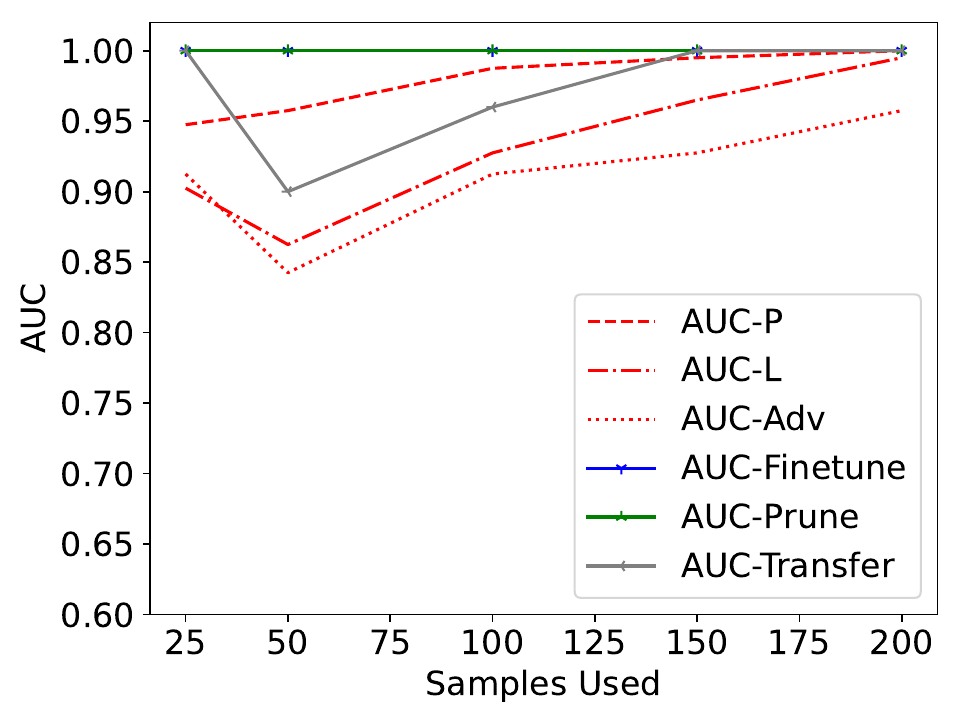} & 
			\includegraphics[width=0.32\linewidth, clip]{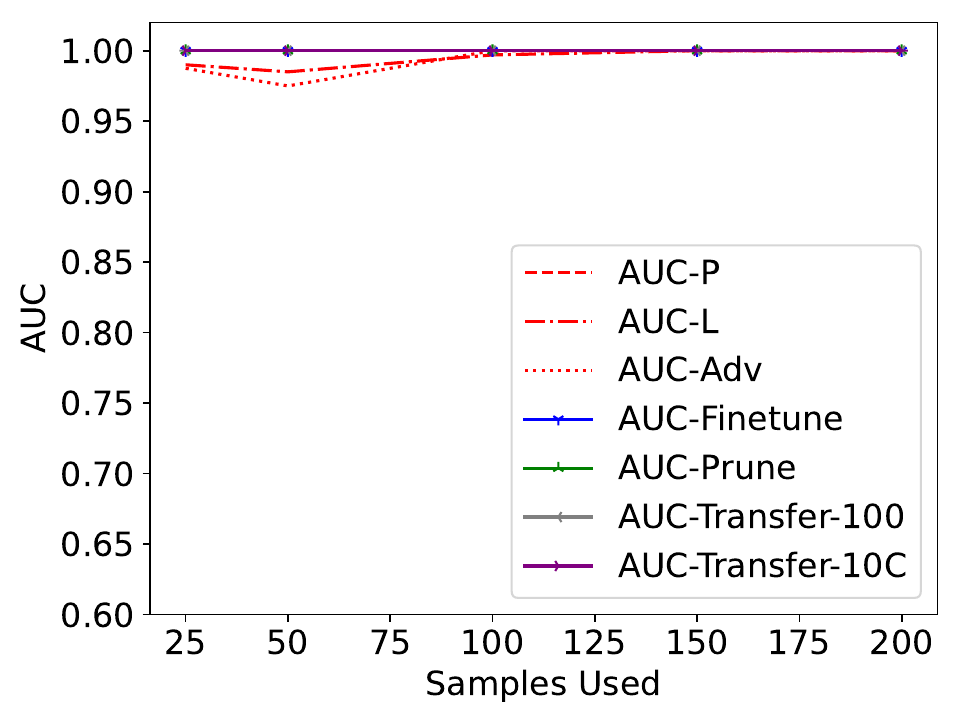} \\
			\\[0.5mm]
			(a) KDEF. & (b) Tiny-ImageNet.  & (c) CIFAR10
		\end{tabular}
        \vspace{1pt}
		\caption{Performance change of SAC-JC with different data amounts.}
		\label{fig:auc_sample_num}
		% \vspace{-10pt}
\end{figure*}

\begin{table}[h]\footnotesize
 \caption{Threshold of SAC-JC on different model architectures on KDEF.}
  \centering  
   \setlength{\tabcolsep}{1mm}{
    \begin{tabular}{ c  c  c  c  c  }  
    \hline  
    {}&  {VGG}&  {ResNet } &  {MobleNet} & {DenseNet} \cr\hline
    {Threshold} & $0.321$ & $0.297$ & $0.295$ & $0.301$\cr\hline
  
    \end{tabular}  
    }
 
    \label{tab:threshold}  
\end{table}

\subsection{Influence of Fingerprinting on the Defender}\label{sec:influence}
When a defender aims to implement model fingerprinting or watermarking, there is typically a trade-off involved, often resulting in certain sacrifices, such as increased time consumption or a decrease in the source model's accuracy.
In Table \ref{tab:time_consumption}, we present the time consumption and the impact on the source model of various model IP protection methods. 
Firstly, we address the issue of accuracy decline. 
Among these methods, only EWE leads to a reduction in the source model's accuracy. 
This is because model fingerprinting techniques, including IPGuard, CAE, and SAC-JC, do not require changes in the training process of the source model, thereby preserving the accuracy of the source model, and on the contrary, EWE necessitates training the source model on watermarked data, resulting in a notable $4.0\%$ accuracy decrease on CIFAR10.
In terms of time consumption, CAE demands several hours for surrogate model training and adversarial example generation, imposing a significant computational burden on defenders. Furthermore, on larger datasets like ImageNet, CAE requires even more time to construct the fingerprint.
Conversely, SAC-JC only takes 0.16 seconds to generate a model fingerprint, representing a remarkable 34393-fold reduction in time compared to CAE.

\begin{table}[h]\footnotesize
 \caption{Time consumption and source model's accuracy decline.}
  \centering  
   \setlength{\tabcolsep}{1mm}{
    \begin{tabular}{c c c c c  }  
    \hline  
    {}&  {IPGuard \cite{cao2021ipguard}}&  {CAE \cite{lukas2020deep}} &  {EWE \cite{jia2021entangled}} & {SAC-JC} \cr\hline
    {Time} & $456.45$s & $25,536.89$s & $277.31$s & $0.16$s\cr
    {ACC ($\%$}) & $87.3$ & $87.3$ & $83.3(-4.0)$ & $87.3$\cr\hline

    \end{tabular}  
    }
 
    \label{tab:time_consumption}  
\end{table}

\subsection{Ablation Study}

\begin{table}[h]\footnotesize
 \caption{Ablation study on KDEF.}
  \centering
    \begin{tabular}{c c c c c  }  
    \hline  
   {Attack(AUC$\uparrow$)}&  
    {SAC-Clean}&{SAC-CAE}& {SAC-JC}  \cr
    % \cr\cline{2-11}   \rowcolor{white}
    \hline  
    Finetune-A & $1.00$   & $1.00$ &   $1.00$   \cr
    Finetune-L & $1.00$   & $1.00$   &  $1.00$ \cr
    Pruning & $0.92$   & $1.00$  &   $0.99$    \cr
    Extract-L & $0.86$   & $0.86$ &  $0.92$   \cr
    Extract-P & $0.98$  & $0.80$  &   $0.97$  \cr
    Extract-Adv & $0.70$   & $0.71$    &  $0.93$   \cr\hline
    \textbf{Average} & $0.91$  & $0.90$   &  $0.97$  \cr\hline
    
    \end{tabular}  
    \label{tab:ablation_study_kdef}  
\end{table} 

\begin{table}[h]\footnotesize
 \caption{Ablation study on face verification.}
  \centering
    \begin{tabular}{c c c c c  }  
    \hline  
   {Attack(AUC$\uparrow$)}&  
    {SAC-Clean}&{WP-JC}& {SAC-JC}  \cr
    % \cr\cline{2-11}   \rowcolor{white}
    \hline  
    Finetune-A & $1.00$   & $0.00$ &   $0.99$   \cr
    Finetune-L & $1.00$   & $1.00$   &  $1.00$ \cr
    Pruning & $0.97$   & $0.38$  &   $1.00$    \cr
    KD & $0.56$   & $0.81$ &  $0.92$   \cr
    Adv-Train & $1.00$  & $0.02$  &   $1.00$  \cr\hline
    \textbf{Average} & $0.91$  & $0.44$   &  $0.98$  \cr\hline
    
    \end{tabular}  
    \label{tab:ablation_study_fv}  
\end{table}

\begin{table}\footnotesize
 \caption{Correlation function's influence on model fingerprinting.}
  \centering  
  \setlength{\tabcolsep}{1mm}{
    \begin{tabular}{c c c c c c   }  
    \hline  
    {AUC$\uparrow$ (KDEF)}&  {FT} &  {Pruning} & {Ex-L} & {Ex-P} & {Ex-Adv}\cr\hline
    {Cosine} & $1.00$ & $0.99$ & $0.92$& $0.97$& $0.93$ \cr
    {Gaussian} & $1.00$ & $0.97$ & $0.94$& $0.95$& $0.98$\cr\hline
    {AUC$\uparrow$ (FV)} &  {FT} &  {Pruning} & {KD} & {Adv} \cr\hline
    {Cosine} &  $1.00$ & $1.00$ & $0.92$ & $1.00$ \cr
    {Gaussian} & $1.00$  & $1.00$ & $0.91$ & $1.00$ \cr\hline

    \end{tabular}  
 }
    \label{tab:cor_func}  
\end{table}

In this section, we thoroughly examine the selection of input samples, and correlation criteria to assess the effectiveness of SAC-JC.
We choose two tasks including face emotion recognition KDEF and face verification (FV) in face recognition as representatives of the multi-classification tasks and verification tasks and our results are illustrated in Tables \ref{tab:ablation_study_kdef} and \ref{tab:ablation_study_fv}.
In these tables, SAC-Clean represents fingerprinting with SAC using clean samples from the defender's dataset, SAC-CAE represents fingerprinting with SAC using the adversarial examples generated in CAE, and WP-JC represents fingerprinting with the accuracy on the wrongly classified samples from the source model.
Our experiments reveal that utilizing JPEG compressed images as the model input for calculating the model's specific correlation matrix yields superior results compared to using clean samples or adversarial examples as inputs.
% Additionally, the results of SAC-JC compared to WP-JC demonstrate the superior performance of fingerprinting with correlation to fingerprinting the instance-level accuracy
Moreover, as depicted in Table \ref{tab:ablation_study_fv}, utilizing the source model's inaccurately classified JPEG compressed images as model input and leveraging the accuracy of suspect models as a point-wise indicator for model fingerprinting proves to be ineffective.
In addition, SAC-Clean outperforms both CAE and IPGuard in terms of AUC for both the face emotion recognition and face verification tasks.
This highlights the superior performance of fingerprinting with correlation compared to fingerprinting with point-wise accuracy.

Furthermore, we also investigate the impact of different correlation functions for model fingerprinting on face emotion recognition KDEF and face verification (FV).
As shown in Table \ref{tab:cor_func}, both the Gaussian RBF kernel and Cosine similarity effectively fingerprint the source model, with an AUC near $1$. 
In the table, FT, Ex, and Adv are abbreviations for Finetune, Extract, and Adv-Train.
Due to the similar performance of the Gaussian RBF kernel and Cosine similarity, we have opted for Cosine similarity as our correlation function in the aforementioned experiments.

\section{Conclusion}

In this paper, we disclose the threat posed by model stealing attacks in deep face recognition.
To protect the well-trained source models, we delve into model fingerprinting in face recognition and propose a correlation-based model fingerprinting framework SAC. 
To weaken the influence of common knowledge shared by the similar-task models, we propose SAC-JC, which leverages a common data-augmented method, JPEG compression on the images for fingerprinting. 
Furthermore, to overcome the lack of output labels in face verification, we propose FRI to generate specific features of images from the 0-1 verification results.
In contrast to existing adversarial-example-based model fingerprinting methods, SAC-JC utilizes augmented clean samples and excels in two tasks where existing methods fall short: adversarial training and transfer learning.
Additionally, SAC-JC eliminates the need for training surrogate models as well as generating adversarial examples, resulting in significantly faster processing speed compared to other fingerprinting methods (approximately 34,393 times faster than CAE).
% On both verification tasks such as face verification and multi-classification tasks such as KDEF, SAC-JC performs well across different model architectures and against different model stealing attacks.
Extensive results validate that SAC effectively defends against various model stealing attacks in deep face recognition. 
This includes both face verification and face emotion recognition tasks, where it consistently exhibits the highest performance in terms of AUC, p-value, and F1 score.
Furthermore, we extend our evaluation of SAC-JC to object recognition datasets, including Tiny-ImageNet and CIFAR10. 
The results also demonstrate the superior performance of SAC-JC compared to previous methods.

\textbf{Data Availability Statement.}
All the datasets used in this paper are available online. 
MS1MV2~\footnote{\url{https://github.com/deepinsight/insightface/tree/master/recognition/_datasets_}}, 
CASIA-Webface~\footnote{\url{https://github.com/deepinsight/insightface/tree/master/recognition/_datasets_}}, 
KDEF~\footnote{\url{https://kdef.se}}, 
CIFAR10~\footnote{\url{https://github.com/wichtounet/cifar-10}}, 
Tiny-ImageNet~\footnote{\url{https://github.com/DennisHanyuanXu/Tiny-ImageNet}}
can be downloaded from their websites accordingly. 

\section{Acknowledgement}

The authors would like to thank the reviewers and
the associate editor for their valuable comments.

% \newpage
{
% \small
% \bibliographystyle{spbasic}
%\bibliographystyle{unsrt}
%\bibliographystyle{plainnat}
\bibliographystyle{unsrtnat}% for \citet
\bibliography{reference}
}
\end{document}